# COLLIDE-PRED: PREDICTION OF ON-ROAD COLLISIONS FROM SURVEILLANCE VIDEOS

*Deesha Chavan, Dev Saad and Debarati B. Chakraborty*

Dept. of Computer Science and Engg., Indian Institute of Technology Jodhpur,
Jodhpur-342037, India

**ABSTRACT**

Predicting on-road abnormalities such as road accidents or traffic violations is a challenging task in traffic surveillance. If such predictions can be done in advance, many damages can be controlled. Here in our wok, we tried to formulate a solution for automated collision prediction in traffic surveillance videos with computer vision and deep networks. It involves object detection, tracking, trajectory estimation, and collision prediction. We propose an end-to-end collision prediction system, named as 'Collide-Pred', that intelligently integrates the information of past and future trajectories of moving objects to predict collisions in videos. It is a pipeline that starts with object detection, which is used for object tracking, and then trajectory prediction is performed which concludes by collision detection. The probable place of collision, and the objects those may cause the collision, both can be identified correctly with Collide-Pred. The proposed method is experimentally validated with a number of different videos and proves to be effective in identifying accident in advance.

## 1. INTRODUCTION

A sum of 151,113 individuals were killed in 480,652 road accidents across India in 2019, a normal of 414 per day or 17 per hour. By predicting collisions beforehand, lots of them can be prevented, and also call for support can be placed in case any of the accidents actually happens. Automated prediction of the collision mounted with a traffic surveillance system could be very effective to address such issues. Here we tried to come up with such a solution. The objective of the proposed work is shown with an example in Fig.1. That is, the probable objects and place of accident could be predicted with trajectory estimation and object detection and according alarm could be generated.

To this end we propose a pipeline to predict collision using Transformers [7, 12, 14], YOLOv4 [13] and siamRPN [11]. The entire proposed method is graphically summarized in Fig.2 in the form of a block-diagram. Here it is shown that the initial input frame of the video scene is fed to the YOLOv4 network to identify the objects present there in the scene. The moving objects, thus identified by YOLO v4 are then tracked with SiamRPN network and their trajectory is being fed to the Transformer to generate the estimation of future trajectory. The information of different previous trajectory and predicted trajectory is then intelligently integrated with collide-pred method, defined here in this article.

The rest of the article is organized as follows. A brief discussion on related work is provided in Sec. 2. Sec. 3 describes the different segments of collide-pred in details. The algorithm of Collide-Pred is described in Sec. 4. The merits of of this method are demonstrated experimentally along with qualitative and quantitative results, and with suitable comparisons in Sec. 5. Sec. 6 draws the overall conclusions on the work along with its future scopes.

## 2. RELATED WORK

This problem of predicting future trajectories of people has been long investigated and a survey for the same is done in [3], [4]. Early approaches used regression models [20], time-series models [5], etc. Then seq2seq models like large-short-term-memory (LSTMS) and recurrent neural networks (RNNs) [6] trained with large amounts of data also seemed to improve the results further given their capability to incorporate memory. LSTM is also used for trajectory prediction in highways [21]. In our approach we use the advanced transformer architecture (TR) [7] for the same since we argue that the Transformer can better model sequences contextually due to self-attention leading to better predictions. Various frameworks also focus on modelling human-human interaction like game-theory [8], GNNs [9], etc. Then models like [10] (SocialPool) use CNNs to aggregate information from various pedestrians. Liang *et al.* [20] came up with a solution of where the future activities got predicted along with its location. The work of Yao *et al.* [19] is the closest work of our proposed one, where a method for traffic accident prediction was formulated with first person view.

The proposed work is different from all the previous work, since here we aim to develop a solution of accident prediction with surveillance cameras. Moreover, we also aim to detect the classify the objects those are causing the accidents with their characteristics.

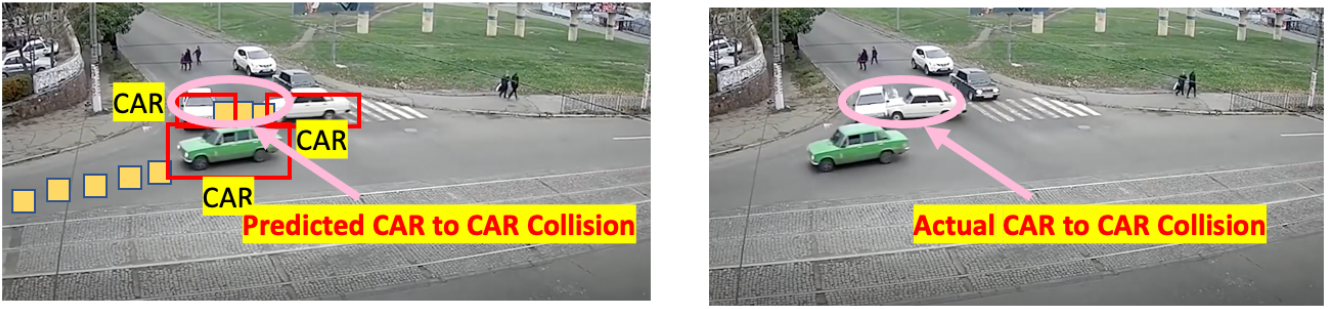

Fig. 1: Visual Representation of Collide-Pred

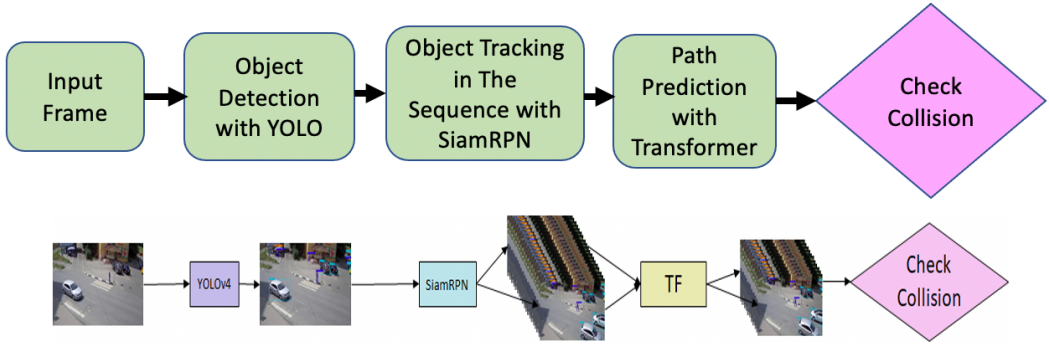

Fig.2: Block Diagram of Collision Prediction with Collide-Pred

## 3. END TO END COLLISION PREDICTION

The details of the method are described here in the section. In other words, the underlying principles of each block in Fig. (1) are described in the rest of the section.

### 3.1. Object Detection

YOLO [13] stands for You Only Look Once. It's an object detection model used in deep learning use case. The first frame yolo finds the moving and static objects present there in the video scene. Let there be $n$ no. of moving objects present there in the scene. These objects will be further tracked, as those are the only object(s) which could cause a collision, may be with another moving object, may be with a static object. Therefore, the objects extracted from the initial input frame $f_0$ are as shown in Eqn. (1).

$$YOLO(f_0) = \{O_0, O_1, ..., O_n\} \qquad (1)$$

Please note, the frames, where some new object will appear, will again be fed to YOLO to detect the new object in the scene.

### 3.2. Object Tracking

The next part of collide-pred is tracking of the moving objects. For the sake of tracking we used the method, SiamRPN [11]. The moving objects, identified by YOLO are then given as input to SiamRPN, where the trcking of n individual object(s) is carried out. Therefore, we are going to have $n$ number of different trajectories ($\tau$), over $P$ no. of previous frames, as the output from this block for a certain frame $f_i$. It is shown in Eqn. (2).

$$SiamRPN(\tau_i) = \{\tau_{i0}, \tau_{i1}, ...., \tau_{in}\} \qquad (2)$$

### 3.3. Object Trajectory Prediction

The trajectories of the moving objects are then fed as the input to the transformer to extract out the estimated trajectory ($\hat{\tau}$) over $Q$ no. of future frames. Therefore, the outcome of this block is $n$ no. of different trajectories as well. The only difference from the previous one is here future trajectory information will be available. The outcome is expressed here in Eqn. (3).

$$TR(\hat{\tau}_i) = \{\widehat{\tau_{i0}}, \widehat{\tau_{i1}}, ..., \widehat{\tau_{in}}\} \qquad (3)$$

### 3.4. Collision Prediction

This is most crucial part of this work, where we have come up with a solution. As we know the future trajectories of the moving objects are the input in this block. Each of the $\widehat{\tau_{ij}}: j = 1, .., n$ consumes $Q$ no. of different locations. We derived a solution of judiciously combining the information of past and future trajectories for decision making. As we can say, if there is a huge variation in the true location of the object than its predicted value, it could be treated as a sudden change in motion, which in return increases the probability of accident/ collision. Let the future trajectory of the $j^{th}$ object be

$$\widehat{\tau_{ij}} = \{\widehat{O_{j1}}, \widehat{O_{j2}}, ..., \widehat{O_{jQ}}\} \quad (4)$$

In Eqn (4) each $\widehat{O_{jl}}$ represents the location of the $j^{th}$ object in the $(i+l)^{th}$ frame and $\widehat{\tau_{ij}}$ is the set of all such locations.. Similarly, each element of the set $SiamRPN(\tau_i)$ in Eqn. (2) contains $P$ discrete locations on P no. of previous frames of each object. Therefore, $\tau_{ij}$ from Eqn. (2) can be written similarly as of Eqn. (4) as follows.

$$\tau_{ij} = \{O_{j1}, O_{j2}, ..., O_{jP}\} \quad (5)$$

Please note that, the first element $\widehat{\tau_{(i+q)j}}$, that is predicted trajectory of $j^{th}$ object in the $(i+q)^{th}$ frame, will be the first element of $\tau_{(i+q+1)j}$, that is, future now, will become past after the next instance of the future. We computed the difference ($\Delta_j$) between the predicted path (Eqn. (4)) and actual path (Eqn. (5)) according to Eqn. (6).

$$\Delta_j = |O_{(i+Q-m)j} - \widehat{O_{(i+m)j}}| : m = 1, ..., Q \ if \ Q \leq P \quad (6)$$

If the values of most of the elements in the set $\Delta_j$ lies above the value of mean($\Delta_j$), it is assumed that $j^{th}$ objet may cause and accident and it is checked further as described below.
There are $n$ no. of different objects present there in the scene, therefore, according to the method, there will be no collision cause by the $j^{th}$ object, if $\widehat{\tau_{ij}}$ satisfies the following condition given in Eqn. (7).

$$\widehat{\tau_{ij}} \cap \widehat{\tau_{ik}} = \emptyset; where \ k = 1, ..., Q \ and \ j \neq k \quad (7)$$

The algorithm of Collide-Pred is described in the next section.

## 4. ALGORITHM

```
Input : Sequence of frames for video
Output : Time segment where collision
occurred

while True:
    if mode == "detection":
        for frame in video:
            object_ids, coordinate =
YOLO(frame)
            check()
    else if mode == "tracker":
        for frame in video:
            if first_frame:
                tracker.init(frame)
            else:
                tracker.update(frame)
                object_ids, coordinate =
tracker.get(frame)
            check()

check():
    history_coordinates <-
append(coordinate)
    if len(history_coordinates) % T ==
0:
        future_coordinates =
Transformer(history_coordinates)

        if intersection between
object_ids and future_coordinates :
            alert(COLLISION
PREDICTED!!)
        else:
            pass
```

## 5. EXPERIMENTAL RESULTS

The proposed method, Collide-Pred is experimentally validated over 30 different videos, that consumes more than 9000 frames. Our method proves to be efficient to successfully predict the accidents and to detect the objects causing causing the accidents. To limit the size of the article we have shown here the outcome of eight such videos, obtained mainly from YouTube [15, 16, 17]. The videos are named as v1,…,v8. It contains different types of videos captured in daylight and artificial light. The videos contain collision between car-to-car (v1 ,v3, v5, v6, v8), car to bus collision (v2) and car to pedestrian collision (v4, v7). All the collisions are successfully detected by our method and it is shown in Fig. 3. The video output video along with codes of the method could be seen in the link [18].

TABLE-I: Efficiency and Accuracy of Collide-Pred Over Different Videos

| S.No | Video | Time-in-Advance (in sec) | Frames in Advance | FP (%) |
|---|---|---|---|---|
| 1 | V1 | 0.56 | 17 | 12 |
| 2 | V2 | 0.9 | 27 | 15 |
| 3 | V3 | 0.35 | 11 | 34 |
| 4 | V4 | 0.5 | 15 | 45 |
| 5 | V5 | 0.42 | 13 | 32 |
| 6 | V6 | 0.3 | 9 | 12 |
| 7 | V7 | 0.12 | 4 | 17 |
| 8 | V8 | 0.23 | 7 | 43 |

Table 1 summarized the results of collision prediction time for those 8 videos. Metric used is the difference between predicted timestamp and actual timestamp (GT) of collision, that is the method can predict the collision in advance of the time listed in the third column of Table I, and the frames in advance are listed in the fourth column. The variation is mainly due the visibilities of the object(s) that changed depending on the positioning of the cameras. Apart from those, the fifth summarized false positive (FP) prediction of collsion over these videos. Please note the rate of generation of false positive alarm is higher if there is major occlusion between different objects.

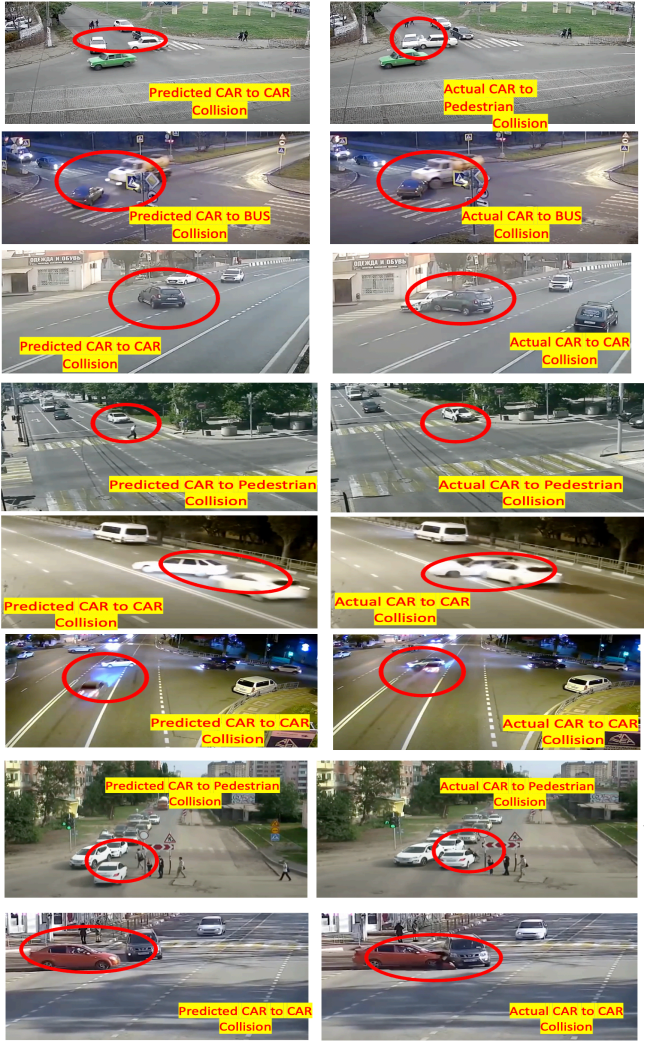

Fig 3: Visual Results of Collide-Pred method on videos v1, v2,.. v8.

### 5.1 Comparative Studies

Since there does not exist any work in the literature with satisfies the same goals as of ours, it won't be fair to carry out a comparative study for our work. However, here we have listed the basic difference between Collide-Pred and the two most closely related work peeking into future (PIF) [20] and road accident prediction (RAP) [19] in Table II.

TABLE-II: Feature-wise Difference Among PIF, RAP and Collide-Pred

| Method | Type of Videos Method is Designed For | Goal |
|---|---|---|
| PIF [20] | Surveillance Videos | Predicting human activities along-with location |
| RAP [20] | Ego-centric Videos | Predicting road accidents being itself a possible object |
| Collide-Pred | Surveillance Videos | Predicting Road Accident along with its location and objects of accident from surveillance videos |

Apart from this, we have also compared trajectory prediction of transformer (TR) [7] with that of LSTM [21]. We replaced the forth block of Fig. 1 in Collide-Pred with LSTM and the accuracy of the observed outcomes are listed in the following Table III in terms of false positive (FP) collision prediction and in advance prediction time (in seconds). It can be seen from the table that although LSTM produced less no. of FP detection, but the in advance prediction time is quite low for it.

TABLE-III: Performance Comparison of Collide-Pred with Transformer and LSTM

| S.No | Video | Tranformer | | LSTM | |
|---|---|---|---|---|---|
| | | FP (%) | Time in Advance (sec) | FP (%) | Time in Advance (sec) |
| 1 | V1 | 12 | 0.56 | 9 | 0.22 |
| 2 | V2 | 15 | 0.9 | 11 | 0.56 |
| 3 | V3 | 34 | 0.35 | 26 | 0.13 |
| 4 | V4 | 45 | 0.5 | 31 | 0.2 |
| 5 | V5 | 32 | 0.42 | 19 | 0.21 |
| 6 | V6 | 12 | 0.3 | 9 | 0.13 |
| 7 | V7 | 17 | 0.12 | 14 | 0.02 |
| 8 | V8 | 43 | 0.23 | 32 | 0.11 |

### 6. CONCLUSIONS AND FUTURE WORKS

The Collide-Pred method defined here, works very well in identifying collision in the videos captured with surveillance cameras. Most interestingly, there is almost no false negative alarm, that is the method is always able to predict the accident. But, there exist some issues related to time and false positive alarms. The prediction time needs to be more a-prior to prevent an accident, as well there should be less false positive alarms generated. Those issues could be addressed in future.